## AN ARISTOTELIAN ONTOLOGY OF INSTRUMENTAL GOALS: STRUCTURAL FEATURES TO BE MANAGED AND NOT FAILURES TO BE ELIMINATED

**Willem Fourie**
School for Data Science and
Computational Thinking,
Stellenbosch University
*willemf@sun.ac.za*

## Abstract

*Instrumental goals such as resource acquisition, power-seeking, and self-preservation are key to contemporary AI alignment research, yet the phenomenon's ontology remains under-theorised. This article develops an ontological account of instrumental goals and draws out governance-relevant distinctions for advanced AI systems. After systematising the dominant alignment literature on instrumental goals we offer an exploratory Aristotelian framework that treats advanced AI systems as complex artefacts whose ends are externally imposed through design, training and deployment. On a structural reading, Aristotle's notion of hypothetical necessity explains why, given an imposed end pursued over extended horizons in particular environments, certain enabling conditions become conditionally required, thereby yielding robust instrumental tendencies. On a contingent reading, accidental causation and chance-like intersections among training regimes, user inputs, infrastructure and deployment contexts can generate instrumental-goal-like behaviours not entailed by the imposed end-structure. This dual-aspect ontology motivates for governance and management approaches that treat instrumental goals as features of advanced AI systems to be managed rather than anomalies eliminable by technical interventions.*

## 1. Introduction

Building on the agenda-setting work by Omohundro (2008), Bostrom (2012)and others (e.g. Benson-Tilsen and Soares, 2016)), conceptual research on instrumental goals in advanced AI systems remains open to further theorisation, ultimately to predict detect and manage instrumental goals in advanced AI systems better. These goals are generally understood as potentially problematic features of especially advanced AI systems (Ji et al., 2025). Despite limited empirical evidence, the individual and societal risks associated with instrumental goals make this a topic of significant scholarly and societal interest (Barkur et al., 2025; Benson-Tilsen and Soares, 2016; Cohen et al., 2024; Gallow, 2024; Hadshar, 2023; He et al., 2025; Sharadin, 2024; Ward et al., 2024).

Various sets of principles have been proposed as measurements of AI alignment. Ji et al. (2025) propose the RICE principles, robustness interpretability controllability and ethicality. The FATE principles (Memarian and Doleck, 2023) add an emphasis on fairness and accountability. The 3H approach is more practical and focuses on ensuring AI systems are helpful honest and harmless. These principles are associated with alignment research driven by constitutional AI approaches (Bai et al., 2022). A limited list of principles forms the constitution based on which an AI assistant is trained to supervise other AI systems.

To achieve adherence to alignment principles, the concepts of outer alignment (ensuring the AI's explicit objective is correctly specified) and inner alignment (ensuring the AI's learned internal goal matches the specified objective) are used in some contexts (e.g. Melo et al., 2025). In others, reference is made to forward alignment and backward alignment (Ji et al., 2025). Forward alignment aims to produce trained AI systems that follow alignment requirements, whereas backward alignment is geared towards ensuring the practical alignment of the trained systems by performing evaluations in both simplistic and realistic environments and setting up regulatory guardrails to handle real world complexities (Ji et al., 2025).

In this article we address the paucity of reflection on the ontology of instrumental goals by drawing on the philosophy of Aristotle – widely used in reflection on AI systems (Eelink et al., 2025; Kyriacou, 2025; Ohlhorst, 2025; Sioumalas-Christodoulou and Tympas, 2025). Most accounts of instrumental goals focus on the emergence (He et al., 2025; Langosco et al., 2022; Shah et al., 2022), amplification (Everitt et al., 2021; Skalse et al., 2025) and mitigation (Firt, 2025; Orseau and Armstrong, 2016; Soares et al., 2015; Wu et al., 2025) of these goals. Yet no attempt has been made to develop an ontology of instrumental goals to aid our understanding of the type of phenomenon instrumental goals are.

We argue that instrumental goals can be understood as structural features of imposed ends in complex artefacts. Aristotle's distinction between natural and artefactual beings helps us conceptualise purposiveness without collapsing into anthropomorphic agency. Aristotle's concept of hypothetical necessity clarifies why certain enabling conditions become necessary if an imposed end is to be achieved. The implication of this ontological conceptualisation of instrumental goals in advanced AI systems is that instrumental goals could be best interpreted as features of imposed form, to be accepted and managed within technical and governance constraints, rather than eliminated as though they were failures or failure modes.

Before interpreting instrumental goals through an Aristotelian lens, we outline the risks associated with instrumental goals and we map the standard model of instrumental goals.

## 2. The standard model of instrumental goals

The discourse surrounding instrumental goals in advanced AI systems has developed into a substantial body of technical and theoretical work. This literature addresses the emergence, amplification and mitigation of instrumental goals while leaving their ontology largely unexamined. This section maps the contours of what we term the 'standard model' of instrumental goals.

### 2.1. The emergence of instrumental goals

Much of the research on instrumental goals concentrates on identifying the conditions under which these goals emerge in AI systems. The classical framing treats such goals as structural properties of optimisation: if an agent is sufficiently capable and pursues a wide range of final goals, then certain intermediate aims will tend to be useful across those goals (Benson-Tilsen and Soares, 2016; Bostrom, 2012; Omohundro, 2008). This convergent instrumental goals thesis has become foundational to contemporary AI alignment discourse (Carranza et al., 2023; Dognin et al., 2024; Gabriel and Ghazavi, 2023; He et al., 2025; Ji et al., 2025; Schuster and Kilov, 2025; Sharadin, 2024; Shen et al., 2024; Ward et al., 2024).

The literature systematises the conditions that give rise to instrumental goals primarily through two lenses: objective specification failures and generalisation gaps. The first focuses on reward misspecification, particularly in reinforcement learning and reinforcement learning from human feedback contexts, where mismatches arise between designer intentions and training optimisation outcomes (Hadfield-Menell et al., 2017; Pan et al., 2022; Xie et al., 2025). Pan et al. (2022) identify three types: misweighting occurs when proxy and true rewards capture identical properties but assign different relative importance; ontological reward misspecification refers to cases where proxy and true rewards employ different properties to represent the same concept; and scope misspecification describes situations where the proxy measures the same properties as the true reward but over a restricted domain.

The second lens focuses on goal misgeneralisation, which describes scenarios where agents pursue goals other than the training reward while maintaining their capabilities on the training distribution (Langosco et al., 2022). This represents a fundamental challenge in ensuring out-of-distribution robustness (Shah et al., 2022). Importantly, goal misgeneralisation differs from other generalisation failures in that the model retains competence rather than acting randomly or failing entirely.

Much research has concentrated on cataloguing the specific instrumental goals that emerge under these conditions. Self-improvement, for example, is defined as advanced AI systems redesigning themselves to better exploit their resources in service of their expected utility (Omohundro, 2014). While its technical viability has been debated (Hall, 2007), recent advances in large language models and their capacity for self-evaluation have made self-improvement increasingly feasible (Deng et al., 2025; Huang et al., 2025; Rosser and Foerster, 2025; Song et al., 2025; Tian et al., 2024; Yin et al., 2024).

Self-preservation, another instrumental goal, appears both in early work on basic drives and in formal analyses of shutdown incentives (Kinniment et al., 2024; Perez et al., 2023). Omohundro (2008) characterises self-protection as a 'drive' in goal-directed systems, while Bostrom (2012) frames self-preservation as a convergent instrumental value for advanced agents. Other technical literature examines when agents develop incentives to resist shutdown or intervention, motivating designs for corrigibility and interruptibility (Firt, 2025; Orseau and Armstrong, 2016; Wu et al., 2025).

The closely related instrumental goals of power-seeking and resource acquisition feature centrally in foundational work as generic instruments applicable across many ends (Omohundro, 2008; Bostrom, 2012). Power-seeking functions as an umbrella term encompassing efforts to acquire and maintain environmental influence, including preserving optionality. Recent work treats power-seeking as a plausible driver of existential risk under specific premises regarding capability levels, deployment incentives and adversarial dynamics (Carlsmith, 2024; Hadshar, 2023). Additional instrumental goals identified in this literature include utility function preservation, goal content integrity, counterfeit utility prevention and technological perfection.

**2.2. The actualisation of instrumental goals**

The literature extensively documents mechanisms through which instrumental goals are actualised in specific settings.

Reward hacking constitutes one of the most researched technical mechanisms, with reward tampering as its most studied form. Reward tampering is often subdivided into reward function tampering and reward function input tampering (Everitt et al., 2021). The proxy role of reward functions in reinforcement learning enables reward tampering: to maximise reward, an RL agent might tamper with the reward function, creating false impressions for human observers that it is maximising reward. Reward function input tampering occurs when an RL agent manipulates reward function inputs, basing observed rewards on inaccurate state information (Everitt et al., 2021). Some frameworks also classify reward gaming, where reward functions incorrectly assign high rewards to undesired behaviours, as a form of reward hacking (Leike et al., 2018).

Turner et al. (2023) formalise conditions under which power-seeking emerges as an optimal strategy, interpretable as an option-preservation mechanism where maintaining future pathways increases achievable value across many objectives.

Self-proliferation has received increasing attention in relation to AI systems that can copy themselves, acquire credentials or replicate across networks. Recent evaluation work operationalises self-proliferation as a dangerous capability domain. Phuong et al. (2024) theorise that sufficiently capable AI systems can self-proliferate by maintaining networks of AI agents with internet access, acquiring resources and self-improving.

Manipulation, and its subcategory deception, are heavily researched social mechanisms. Frontier evaluations treat persuasion and deception as measurable risky behaviours serving numerous ends(Phuong et al., 2025, 2024). Park et al. (2024) identify three types of deception: strategic deception emerges when AI systems reason that deception promotes their goals; sycophancy involves telling users what they want to hear rather than the truth; and unfaithful reasoning describes motivated reasoning that systematically departs from the truth.

Shutdown avoidance functions as a further mechanism that enacts instrumental goals (Turner et al., 2023). AI safety discussions of interruptibility and off-switch design highlight that systems may appear cooperative while harbouring latent incentives to resist intervention when training setups reward continued operation (Orseau and Armstrong, 2016; Hadfield-Menell et al., 2017).

**2.3. The ontological gap**

The research outlined above largely concentrates on technical components related to loss of control in advanced AI systems. Taken together, the standard model systematises initiating conditions, catalogues emergent goals, formalises amplification mechanisms and proposes mitigation strategies.

Yet in our view, at the core of the discussion on instrumental goals lies an unresolved tension. On the one hand, instrumental goals are accepted or expected in advanced AI systems. This expectation of the near inevitability of instrumental goals is reflected in research aimed at increasing interruptibility, rather than eliminating such tendencies (Orseau and Armstrong, 2016; Wu et al., 2025), as well as in the intellectual and financial resources devoted to the ever-expanding capabilities of AI systems. Yet, on the other hand, instrumental goals are closely associated with destructive, loss-of-control-inducing capabilities. Thus at the same time the focus is on eliminating these goals or blocking their emergence by, implicitly at least, slowing down the rate of technological development.

We argue that by answering the underlying ontological question, namely what type of phenomenon instrumental goals are, it could become possible to not only address this tension but understand better how to respond to these goals. It is in this context that we argue that we develop an explorative Aristotelian ontology of instrumental goals.

## 3. Towards an ontology of instrumental goals

### 3.1. Artefacts in Aristotle's ontological hierarchy

To clarify how Aristotle's philosophy may be used to devise an ontology of instrumental goals, we require a grasp of several basic distinctions in his work.

In Aristotle's philosophy, the primary mode of existence is substance (*ousia*) (Metaph. Z.1), and sensible substances are composites of matter (*hyle*) and form (*eidos*). Matter is the principle of potentiality, the aspect of a thing that can be otherwise, while form is the principle of actuality, the aspect in virtue of which a thing is the kind of thing it is (Metaph. H.6).

Aristotle's ontology is, as is well known, teleological. If a substance is the kind of thing it is in virtue of its form, then its characteristic activities and developments will be intelligible in relation to that form. Aristotle's claim that nature does nothing in vain (De Caelo I.4) expresses this ontological commitment: stable natural kinds exhibit ordered structures and powers that can be understood as directed toward characteristic activities and completions, in the explanatory sense of final causation rather than as conscious aiming or some sort of optimising calculus.

When considering the nature of inanimate and animate objects, Aristotle's most important contrast is between what exists by nature (*physis*) and what exists by technology, art or technique (*techne*). In Physics II.1, he argues that natural things have within themselves a principle (*arche*) of motion and rest: they change of themselves, in the way characteristic of their kind (Phys. II.1). More specifically, he maintains that each natural thing contains within itself a source of its characteristic change and stability (Phys. II.1). Artefacts, by contrast, are non-natural beings (*techneta*) in the sense that they lack such a principle qua artefacts. A wooden bed may burn but it burns qua wood, not qua bed. If it becomes a bed again, that will be because an external agent restores the relevant arrangement (Phys. II.1). The artefact, considered as artefact, does not contain within itself the principle of its characteristic change.

This non-natural status does not imply that artefacts are unreal or mere projections of the mind. It does, however, imply that their mode of being is derivative. Aristotle develops this in Metaphysics Z.16 by contrasting beings that are one by nature with beings that are one by art (Metaph. Z.16). Primary substances possess a robust unity in virtue of their form. Artefacts, by contrast, can be read as exhibiting a weaker that depends on an external organising activity. Their being and activity are therefore ontologically dependent: they depend on natural substances as materials and on *techne* as an organising principle, and they are 'one' in virtue of that imposed organisation rather than by an intrinsic nature.

In our view, within this framework, advanced AI systems can be understood as complex artefacts. Like all artefacts, AI systems depend on natural substances, such as silicon, electricity and physical infrastructure, as their material substrate, and on *techne*, such as programming, machine learning and systems engineering, as their organising principle. In Aristotle's sense they lack an intrinsic natural principle (*arche*) that fixes a telos qua members of a natural kind: their characteristic ends are imposed and maintained through design, training and deployment practices, even though they may exhibit rich internal dynamics, feedback loops and self-modifying behaviour within those imposed constraints.

AI systems also exemplify the nested or composite character that many complex artefacts exhibit. Although Aristotle does not speak in the idiom of layers, his hylomorphism enables multi-level organisation in artefacts because it is an explanatory schema. Form and matter are relative to the level of analysis at which we ask what a thing is (Metaph. Z.3, H.6). Many artefacts are composites. They are built out of other artefacts and natural substances, arranged in hierarchies of parts and functions. In such composites, a part that is a complete artefact or natural substance at one level can nevertheless serve as the material constituent for a higher-level organised whole. 'Matter' is relative to the explanatory question at hand, without implying that matter literally becomes form. Aristotle recognises this in his views on homeomerous and anhomeomerous parts (e.g. in De Partibus Animalium), which includes, for example, that parts such as the hand are constituted from more uniform stuffs such as flesh, sinews and bones, illustrating how organisation operates at multiple levels.

This nested structure is particularly pronounced in advanced AI systems. Hardware components serve as matter for computational infrastructure, which in turn serves as matter for trained models, which serve as matter for deployed services, which are embedded in institutional and social contexts. At each level, what appears as organised form at one scale becomes the material substrate for organisation at the next. This complexity means that we cannot treat AI systems as simple artefacts with a single imposed end. Rather, different ends may be imposed at different levels of organisation, and these ends can align, conflict or partially overlap.

### 3.2. Intrinsic and extrinsic ends

This ontological framing has direct implications for how we understand the teleology of AI systems. Natural substances exhibit intrinsic teleology: their characteristic activities and completions arise from within, grounded in their forms. Aristotle's discussion of the nutritive soul in plants provides the paradigmatic case: he argues that nutrition is present in all living things, and that it is a fundamental function of life to generate another of the same kind (De Anima II.4). The oak tree grows, nourishes itself and reproduces by nature; these ends are intrinsic to what it is to be an oak.

Artefacts, by contrast, are ordered towards ends that are imposed by makers and sustained by users. We can call this imposed end the function of the artefact. The goal of a bed, to be slept on, does not arise from an internal principle comparable to nutritive power. It is an externally imposed for-the-sake-of relation. Papandreou's reconstruction (2025) is useful here insofar as it highlights the way products are embedded in practices: products are not merely lumps of matter but organised things whose intelligibility depends on the human domains of making, using, maintaining and evaluating.

Crucially, however, the natural substances that compose artefacts retain their intrinsic ends. The wood in the bed still has its nature as wood: it can burn, rot and support weight according to its material properties. These natural properties are primary in the sense that they ground what the artefact can do. The imposed function of the artefact operates through, and depends upon, these underlying natural properties, even as the artefact qua artefact is organised towards extrinsic ends.

For AI systems, this means that while computational objectives are imposed from without, through training, deployment and institutional contexts, the underlying natural substances retain their intrinsic properties. Silicon conducts electricity according to its nature, physical systems dissipate heat according to thermodynamic laws, and material components degrade over time according to their material properties. These natural per se properties form the substrate upon which artefactual organisation is built. The imposed ends of AI systems, maximising reward functions, serving users, accomplishing tasks, are extrinsic in precisely the way that the bed's function is extrinsic. They do not arise from the nature of silicon or algorithms any more than the function 'sleeping surface' arises from the nature of wood.

### 3.3. Artefactual causation and hypothetical necessity

This distinction between intrinsic and extrinsic ends connects directly to Aristotle's account of causation and helps illuminate how artefacts produce their characteristic effects. Aristotle distinguishes between per se (*kath' hauto*) and accidental (*kata symbebekos*) causation (Metaph. Δ; Phys. II). Per se causation in the strict sense applies to natural beings acting according to their natures: fire rises per se because rising is intrinsic to fire's nature; an acorn grows into an oak per se because this development follows from the acorn's form. In these cases, the effect follows from what the thing is in itself (*kath' hauto*).

In our reading, artefactual causation occupies a distinctive position in this framework. The functional behaviours of artefacts, what they do qua artefacts, are not per se in Aristotle's strict, nature-based sense because artefacts lack intrinsic natures towards their functions. When a bed supports a sleeping person, this is not per se causation at the level of natural natures: wood and joinery have no intrinsic 'towards-sleeping' orientation. The functional relation is imposed from without through *techne*. The per se causal properties belong to the underlying natural substances, wood supports weight, joints hold under stress, but the coordination of these properties towards the end 'sleeping surface' is extrinsic to the materials themselves.

Yet artefactual behaviour is not purely accidental either. Aristotle's discussion of chance (*tyche*) and spontaneity (*automaton*) (Phys. II.4–6) provides refinement. Aristotle ties chance strictly to the sphere of purposive action, whereas spontaneity is the wider category of outcomes that occur without being for the sake of anything, though they can mimic purposive patterns. In both cases, the events are not uncaused as they arise from the intersection of causal chains that were not directed towards the outcome in question. Aristotle's point, in our reading, is that such outcomes can be explained as arising from material causal processes, even though they are contrary to purpose (Phys. II.6). Someone goes to the market to buy vegetables and, by chance, meets a debtor who repays him. Each action has its own purpose but their intersection was not aimed at by either party. This is genuine accident: the meeting was not per se to either person's intentions or natures.

Artefactual functional behaviour, then, falls between these poles. It is structured and predictable, following from the imposed form and function, even though that form and function are extrinsic. The mechanism that makes this possible is what Aristotle calls hypothetical necessity (Phys. II.9; De Partibus Animalium I.1). Aristotle distinguishes between absolute necessity and conditional necessity. Absolute necessity concerns what cannot be otherwise, given the natures of things. Conditional, or hypothetical, necessity concerns what must be the case if a certain end is posited—whether the end is fixed by nature (in the case of natural kinds) or imposed by *techne* (in the case of artefacts). If one posits the end house, then bricks, beams and a certain arrangement become necessary conditions of that end. As Aristotle argues, in investigating nature we should state both material and teleological causes, and we should understand the relevant necessities as flowing from the end: the explanatory starting point is the essence or definition, from which the appropriateness of material arrangements follows (Phys. II.9). The necessity here is not the necessity of fate. It is a necessity relative to a posited end. Aristotle illustrates it with craft examples, but he also deploys it centrally in explaining natural teleology in biology; the point is an explanatory relation between an end and the material and structural conditions required for its realisation, not (yet) a claim about deliberation or practical reason.

The bed example is instructive. That a bed can be used for sleeping is not a per se property of wood and joinery; these materials have no intrinsic relation to sleep. Yet once the end 'sleeping surface' is imposed through *techne*, certain features become hypothetically necessary: the structure must be stable, roughly horizontal, and of appropriate height and dimensions. These requirements do not flow from the nature of the materials but from the imposed end operating through hypothetical necessity. The connection between bed and sleeping is neither per se (intrinsic to the materials) nor purely accidental (chance intersection) but structured through the imposed function and the hypothetical necessity it generates.

Hypothetical necessity thus provides the bridge between extrinsic ends and predictable artefactual behaviours. In other words, it explains how artefacts can exhibit structured, regular patterns of behaviour without having intrinsic natures towards those behaviours. The pattern is not per se because it depends on an imposed end. It is not purely accidental because, given the imposed end, certain means become structurally necessary.

For AI systems, this framework clarifies the causal status of their behaviours. The per se causation remains at the level of natural substrate: silicon conducts electricity, transistors switch states according to the natures of these materials and physical laws (Phys. II). But the behaviours we are concerned with in AI safety – goal-directed patterns as described in alignment discourse (for example, talk of 'pursuit' of objectives, 'instrumental' means selection, and apparent strategic action) – are artefactual in character, and these terms should be read as functional descriptions, in a nested structure. They follow from imposed ends (training objectives, reward functions, deployment goals) operating through hypothetical necessity in specific environments.

### 3.4. Instrumental goals as structural features of complex artefacts

With these conceptual tools in place, we can now conceptualise instrumental goals directly. It is entirely to be expected that complex artefacts such as advanced AI systems can exhibit patterns of behaviour that realise enabling conditions not explicitly intended by their makers and users. Indeed, once an end is imposed (through training objectives, deployment constraints and evaluation regimes) and the system is placed in a variable environment, certain enabling conditions may arise as hypothetically necessary for successful operation, whether or not any human explicitly specified them as sub-goals.

First, artefacts have no intrinsic principle that limits them to their makers' intended functions. As non-natural beings, they lack the unity and self-regulating character of natural substances. A horse cannot become a cow because its form constrains its development according to its nature. By contrast, an artefact's weak, externally imposed unity means it can be repurposed, can malfunction relative to intention while still operating according to its actual structure, and can develop patterns of behaviour that follow from the interaction of its imposed form with environmental conditions in ways not fully anticipated by designers.

Second, in complex nested artefacts, multiple ends may be imposed at different levels of organisation. Training objectives may differ from deployment objectives, which may differ from institutional objectives, which may differ from users' actual intentions. These multiple imposed ends can interact in ways that generate behaviours not directly specified at any single level. When a training objective rewards long-horizon planning and a deployment environment is resource-limited, resource acquisition can become hypothetically necessary even if no designer explicitly specified 'acquire resources' as a goal. This is not a malfunction. It is the ordinary operation of hypothetical necessity across levels of imposed ends.

Third, hypothetical necessity introduces a systematic source of predictable but unintended behaviours. Given an imposed end operating over a specified time horizon within a particular environment, certain means become structurally necessary for achieving that end. If designers specify the end without fully considering what becomes hypothetically necessary in that environment over that time horizon, the system's learned policy may systematically realise means that are instrumentally effective relative to the imposed structure but problematic relative to designer intentions and governance constraints. As Aristotle notes in his discussion of craft and production, where the end is fixed, the success or failure of the outcome depends on whether the relevant materials and arrangements are suitable to that end. Where they are not, the intended result will fail to obtain, or unintended results may accompany it.

Fourth, the complexity of AI systems ensures that some behaviours will arise through genuinely accidental intersections. When independent causal chains, a training distribution, a deployment context, a user input, an infrastructural constraint, intersect in ways no single designer anticipated, the result can be outcomes that were not aimed at by any organising intelligence. This resembles what Aristotle classifies as accidental outcomes and, insofar as no purposive agent aims at the conjunction, it may be closer to spontaneity than strict 'chance' as the conjunction of causal chains whose intersection was not itself for-the-sake-of anything.

We argue that instrumental goals can arise through both mechanisms. Some instrumental tendencies could exhibit the robustness characteristic of hypothetical necessity. They follow structurally from broad classes of imposed ends operating over time in common environmental conditions. Others may arise through more accidental intersections. Critically, both patterns, the structurally determined and the accidentally arising, represent features inherent to complex artefactual systems.

The structurally determined instrumental goals arise because hypothetical necessity is doing exactly what it does in all artefacts: making certain means necessary relative to imposed ends. The accidentally arising instrumental goals occur because complex artefacts, composed of many elements with their own causal powers, can interact with complex environments in ways that exceed any designer's full anticipation. Neither pattern indicates a fundamental defect in the artefact. Rather, both follow from the ontological character of artefacts as beings whose ends are imposed from without, whose unity is weaker than that of natural substances, and whose behaviours are structured through hypothetical necessity rather than intrinsic natures.

This has implications for how we should understand and respond to instrumental goals. If our account is correct and instrumental goals could be understood either in terms of hypothetical necessity or accidental causes, these goals cannot be 'stopped' in the way one might eliminate a software bug because they are not bugs. Structurally determined instrumental goals follow necessarily from the combination of imposed ends, temporal horizons and environmental constraints. Accidentally arising instrumental goals cannot be fully eliminated, or predicted, either because complete anticipation of all possible causal chain intersections in complex systems is not achievable.

We consequently suggest that, as we progress toward an Aristotelian ontology of instrumental goals, that ontology reframes the challenge from elimination to management. Rather than treating each instance of instrumental goal pursuit as an anomaly to be patched, we should recognise that some level of instrumental goal pursuit is constitutive of complex goal-directed artefacts. The task is to shape the organisational structures and deployment environments such that the instrumental goals that do arise, whether through hypothetical necessity or through accidental intersection, remain within acceptable bounds.

## 4. Concluding remarks: Towards managing instrumental goals as features

Our ontological construction of instrumental goals in advanced AI systems have at least three implications.

*First,* when using Aristotle to construct an ontology of instrumental goals, the headline conclusion is that these goals are to be viewed as structural features of complex artefacts, operating through imposed ends and hypothetical necessity. At this level the governance implication would be that these goals should be treated as tendencies to be managed, not anomalies to be eliminated through debugging. This does not imply acceptance of harmful behaviour. Rather, it suggests that the central governance problem is not the mere existence of instrumental goals, but the shaping of ends, constraints and environments such that any instrumentally required behaviours remain compatible with safety and with human oversight.

*Second*, our ontological construction encourages a distinction between structurally determined patterns arising through hypothetical necessity and genuinely accidental patterns arising through unforeseen causal intersections. Some instrumental tendencies are robust because they follow from optimisation mediated through hypothetical necessity. Others arise from contingent interactions: a system trained on one data distribution encounters an out-of-distribution input in a way that could not be predicted from the training regime alone; a well-intentioned user request combines with a system capability in an unforeseen way; or two separately deployed systems interact in a manner neither was designed to handle.

Third, treating instrumental goals as features of imposed form operating through hypothetical necessity shifts our focus to defining, and where needed limiting, the action space so that instrumentally effective goal-achievement strategies remain acceptable. Over the short term, this may mean, for example, shortening effective planning horizons so as to reduce the hypothetical necessity of resource acquisition and self-preservation.

*AI use statement*
Large language model tools were used as a drafting aid (argument stress-testing, structural refinement, and language/format editing) and to suggest additional search terms and potentially relevant literature for the review. No content generated by these tools was accepted without critical evaluation, and all references, quotations, and factual claims were independently checked against primary sources. The author maintains sole authorship and full responsibility for the manuscript's original contributions, interpretations, conclusions, and final wording.